\documentclass[lettersize,journal]{IEEEtran}
\usepackage{amsmath,amsfonts}
\usepackage{algorithmic}
\usepackage{algorithm}
\usepackage{array}

\usepackage[caption=false,font=normalsize,labelfont=sf,textfont=sf]{subfig}

\usepackage{textcomp}
\usepackage{stfloats}
\usepackage{amsmath}
\usepackage{amssymb}
\usepackage{url}
\usepackage{verbatim}
\usepackage{graphicx}
\usepackage{multirow}
\usepackage{amsmath}
\usepackage{amsfonts}
\usepackage{algorithm}
\usepackage{algorithmic}
\usepackage{bm}
\usepackage{cite}

\usepackage{tikz}
\usetikzlibrary{positioning, shapes.geometric, backgrounds}
\usepackage{float} %

\begin{document}

\title{MLA-CoE: A Novel Co-Evolution Method for Label-Only Attacks in Multi-Label Classification Models}

%\author{IEEE Publication Technology,~\IEEEmembership{Staff,~IEEE,}
%        % <-this % stops a space
%\thanks{This paper was produced by the IEEE Publication Technology Group. They are in Piscataway, NJ.}% <-this % stops a space
%\thanks{Manuscript received April 19, 2021; revised August 16, 2021.}}

\author{Zhijian~Chen, Yiya~Diao, Yamin~Hu, Yujiang~Liu, Yuhui~Shi,~\IEEEmembership{Fellow,~IEEE} and 
	Wenjian~Luo,~\IEEEmembership{Senior Member,~IEEE}
	\thanks{This study is supported by the National Natural Science Foundation of China (Grant No. 62576121), University Stability Support Program of Shenzhen (Grant No. GXWD20231130113127003), Shenzhen Science and Technology Program (Grant No. ZDSYS20210623091809029).
		(\textit{Corresponding author: Wenjian Luo.})}
	\thanks{Zhijian Chen, Yiya Diao, Yamin~Hu, Yujiang Liu and Wenjian Luo are with the Guangdong Provincial Key Laboratory of Novel Security Intelligence Technologies, Institute of Cyberspace Security, School of Computer Science and Technology, Harbin Institute of Technology, Shenzhen 518055, China.}
	\thanks{Yuhui Shi is with the Department of Computer Science and Engineering, Southern University of Science and Technology, Shenzhen 518055, China.}
	
	\thanks{Email: 21B951010@stu.hit.edu.cn, diaoyiya@hit.edu.cn, huyamin@hit.edu.cn, 23S151125@stu.hit.edu.cn, shiyh@sustech.edu.cn, luowenjian@hit.edu.cn}
}

%\author{\IEEEauthorblockN{Anonymous Authors}}

%% The paper headers


\maketitle

\begin{abstract}
	Multi-label black-box attacks pose severe security threats to multi-label classification models and have attracted widespread attention in the field of adversarial machine learning.
	For constructing black-box attack algorithms, a common and effective strategy is to model the attack as an optimization problem. 
	However, under the extreme scenario where only model output labels are accessible (which is also known as the label-only case), it remains extremely difficult to implement effective attack algorithms against multi-label classification using traditional evolutionary algorithms.
	This difficulty stems from the large problem scale and limited information, which often leads to slow adversarial example generation and high query consumption during optimization.
	To address this challenge, we propose a multi-label label-only attack algorithm, MLA-CoE, based on co-evolution for ``hiding single'' attacks. 
	MLA-CoE adopts a sparsification strategy for dimensionality reduction and uses cooperative co-evolution to decompose the large-scale optimization task into smaller, tractable subproblems, which are then solved by a modified differential evolution algorithm.
	This allows for precise control of the perturbation magnitude and efficient utilization of query counts, resulting in faster and more accurate label-only black-box attacks.
	We conducted experiments on two mainstream multi-label classification models and two multi-label datasets. The results demonstrate that our method not only effectively limits the size of the perturbation but also significantly improves the efficiency of adversarial example generation, validating the effectiveness and potential advantages of cooperative co-evolution in multi-label adversarial attacks. This research contributes to the development of practical methods for generating multi-label label-only black-box attacks and testing the robustness of existing multi-label classification models.
\end{abstract}

\begin{IEEEkeywords}
Deep neural networks, multi-label classification, adversarial examples, cooperative co-evolution.
\end{IEEEkeywords}

\section{Introduction}
In recent years, deep learning models have been widely deployed in safety-critical domains such as autonomous driving~\cite{auto_drive}, medical diagnosis~\cite{medicine}, and facial recognition~\cite{el2025facial}.
However, research confirms that the deep learning models deployed in these systems are inherently vulnerable to adversarial examples, which are carefully crafted inputs with subtle perturbations that can cause the models to make incorrect predictions~\cite{karunanayake2025out}.
The existence of such malicious samples poses severe threats to the reliability and security of intelligent systems, potentially leading to catastrophic real-world consequences~\cite{samek2021explaining,inkawhich2023adversarial}.
Therefore, investigating adversarial attacks is not only critical for exposing the robustness flaws of deep learning models but also essential for developing defensive strategies to enhance the security of real-world AI systems.

Adversarial attack methods are broadly categorized into white-box attacks and black-box attacks based on the attacker’s access to target model information~\cite{han2023interpreting,liang2022adversarial}.
In white-box scenarios, adversaries have full knowledge of the target model (e.g., architecture, weights, and gradient information), enabling the design of highly efficient attack algorithms. Representative white-box techniques include the Fast Gradient Sign Method (FGSM) \cite{fgsm} and Projected Gradient Descent (PGD) \cite{pgd}, which leverage gradient information to compute minimal perturbations that trigger misclassification~\cite{wan2023average}. 
While white-box attacks achieve high success rates with small perturbations, they are impractical in real-world scenarios, where model details are typically proprietary or inaccessible.

In black-box attacks, attackers can only access the model's input-output pairs~\cite{zhu2024review}.
This makes black-box attacks more realistic but also far more challenging: attackers must model the target model’s behavior indirectly and generate adversarial examples through iterative querying of the model’s outputs. 
Currently, formulating black-box attacks as optimization problems is a common and effective approach to generating adversarial examples~\cite{zheng2025blackboxbench}.
In targeted attacks, the optimization objective is to find the minimal perturbation that meets the attack target while satisfying query budget constraints.
Evolutionary algorithms (EAs) have emerged as a common solution for this task—their gradient-free nature and global search capability make them well-suited for black-box optimization, and they have achieved promising results in single-label (multi-class) black-box attack scenarios~\cite{opa,Genattack,nesmachnow2025adversarial,williams2024evolutionary,qiu2021black}.

Despite substantial progress in multi-class black-box attacks, black-box attacks face greater challenges in multi-label classification models, making many existing multi-class black-box attack methods ineffective~\cite{chen2}. 
Unlike multi-class classification models, which assign a single label to each sample, multi-label classification models output multiple labels for each input, and the classification decision is controlled by a specific threshold~\cite{zhang2025three}.
This characteristic typically results in correlations among labels in multi-label classification models.
The inter-label correlations and high-dimensional output space make multi-label adversarial attacks far more complex.
For example, attackers must manipulate target labels while ensuring that other labels remain unaffected and that the perturbations introduced are imperceptible.

However, research on label-only black-box scenarios remains relatively scarce. ML-LBA~\cite{kong2}, the state-of-the-art multi-label black-box attack method, employs a YOLO initialization~\cite{yolov3} to generate initial adversarial examples and then develops ML-BA~\cite{kong2} based on the boundary attack~\cite{ba}. Nevertheless, these methods still suffer from low efficiency and excessive query consumption when handling high-dimensional inputs, and traditional evolutionary algorithms easily become trapped in local optima when treating the perturbation vector as a single unified optimization problem~\cite{lehre2023self}.
To address these challenges, we propose MLA-CoE, a cooperative co-evolution-based multi-label adversarial attack algorithm tailored specifically for the label-only black-box scenario. 
MLA-CoE first introduces a sparsification strategy to reduce the problem dimensionality by retaining only critical perturbation pixels. It then employs a cooperative co-evolution framework to decompose the high-dimensional optimization problem into collaborative low-dimensional subproblems, which are solved using a modified differential evolution (DE) algorithm.
We evaluate MLA-CoE on two mainstream multi-label classification models (ML-GCN~\cite{ML-GCN} and ML-LIW~\cite{song}) and two benchmark datasets (VOC 2007~\cite{VOC2007} and VOC 2012~\cite{VOC2012}). 
Experimental results demonstrate that MLA-CoE outperforms the two state-of-the-art label-only multi-label attack methods by producing higher-quality initial adversarial examples, achieving higher attack success rates, and generating smaller adversarial perturbations. These results validate that cooperative co-evolution algorithms are superior to traditional evolutionary algorithms for multi-label black-box attacks.

In summary, the main contributions of this paper are as follows.
\begin{enumerate}
	\item We propose MLA-CoE, a novel co-evolutionary attack method for the ``hiding single" task in multi-label label-only black-box settings. MLA-CoE decomposes the high-dimensional adversarial example generation task into tractable subproblems, which effectively improves attack efficiency and success rate.
	\item We design a sparsification strategy to retain only critical pixels for adversarial attack, which reduces the optimization dimension and computational overhead. We further present a variable grouping strategy to enhance the convergence speed and stability of the evolutionary algorithm.
	\item Extensive experiments on two mainstream multi-label models and two benchmark datasets show that MLA-CoE outperforms ML-LBA and ML-BA by accelerating adversarial example generation, boosting success rate, and reducing perturbation magnitude. Our method achieves particularly strong performance under limited query budgets.
\end{enumerate}

The remainder of this paper is organized as follows. Section \ref{sec2} introduces the background knowledge related to the work of this paper, mainly including various adversarial attack algorithms, label-only adversarial attacks, and cooperative co-evolutionary algorithms. Section \ref{sec3} describes the algorithm flow of MLA-CoE in detail, including YOLO initialization, cooperative co-evolutionary strategy, and differential evolution. Section \ref{sec4} introduces experimental setup. Section \ref{sec5} presents experimental results and analysis. Section \ref{sec6} briefly summarizes this work.

\section{Related work}
\label{sec2}
In this section, we will first provide a brief overview of existing multi-label adversarial attacks. We will categorize the methods for generating adversarial examples based on the information available to attackers, focusing on multi-label black-box adversarial attacks. Next, we will briefly introduce label-only attacks. Finally, we will discuss the relevant background on co-evolution.

\subsection{Multi-label Adversarial Attacks}
\label{Multi-label Adversarial Attacks}
As mentioned earlier, multi-label adversarial attacks are elaborated from two perspectives: white-box attacks and black-box attacks. In white-box scenarios, Song et al. \cite{song} proposed multi-label variants of four classic multi-class white-box attack algorithms (FGSM \cite{fgsm}, FGM \cite{FGM}, C\&W \cite{cw}, and DeepFool \cite{deepfool}) via transplantation. Additionally, Song et al. designed a ranking-based multi-label attack framework, which includes methods such as Rank \uppercase\expandafter{\romannumeral1} and Rank \uppercase\expandafter{\romannumeral2} \cite{song}. Zhou et al. \cite{mlalp} constructed the MLA-LP attack method dedicated to multi-label classification models by modeling multi-label adversarial attacks as linear programming problems. Chen et al. \cite{chen1} transplanted five classic gradient-based multi-class attack algorithms to the multi-label setting. They further validated the performance differences of different gradient-based attack strategies in multi-label scenarios through extensive experiments.

In the field of black-box attacks, Kong et al. \cite{kong} first proposed MLAE-DE to solve multi-label black-box attacks using differential evolution (DE). MLAE-DE leverages DE to achieve high attack success rates for the ``hiding single'' and ``hiding all''~\cite{luohide} multi-label black-box attacks.
In the algorithm, Kong et al. \cite{kong} introduced a complementary mutation operator to reduce the query cost of attacks, specifically for the ``hiding all'' attack setting.
Furthermore, Chen et al. \cite{chen2} transplanted seven query-based multi-class attack methods to the multi-label scenario, and systematically compared the attack efficiency and perturbation magnitude of these methods on mainstream multi-label datasets.
Besides, Hu et al. \cite{hu2021tkml} proposed TkML-AP, an adversarial attack method targeting top-k multi-label learning, which further extends the application scope of multi-label adversarial attacks to ranking-oriented tasks.

\subsection{Label-only Adversarial Attacks}
\label{Differential Evolution}
Label-only adversarial attacks, also known as decision-based black-box attacks, are the most restrictive class of black-box attacks, where the attacker can only obtain discrete predicted labels from the model without access to confidence scores or internal model information. Unlike score-based attacks that leverage confidence values for optimization, label-only attacks lack numerical gradient feedback and can only judge attack effectiveness through label changes, making them extremely challenging.

Brendel et al.~\cite{ba} the Boundary Attack (BA) \cite{ba}, which is the pioneering work in label-only attacks. Starting from an initial adversarial example, BA iteratively optimizes the example along the model's decision boundary to reduce the perturbation magnitude while maintaining the adversarial label, laying the foundation for subsequent methods.

On the basis of BA, Chen et al.~\cite{hsja} proposed HSJA method, which locates the decision boundary efficiently via binary search and estimates the gradient through finite differences, significantly reducing the number of queries. Li et al.~\cite{qeba} further proposed QEBA method, which projects the perturbation search into a low-dimensional subspace using Monte Carlo sampling to improve gradient estimation efficiency for high-dimensional data. 
In addition, Cheng et al.~\cite{cheng2019sign} proposed Sign-OPT, which simplifies direction estimation by sampling discrete symbol vectors, making it suitable for real-time attack scenarios. Maho et al.~\cite{surfree} proposed SurFree, which achieves efficient boundary search from a geometric perspective without explicit gradient estimation.

However, the methods mentioned above are all designed for multi-class classification models. 
There are relatively few attack algorithms designed specifically for multi-label classification models. 
To the best of our knowledge, the ML-LBA attack algorithm proposed by Kong et al.~\cite{kong2} is the only label-only black-box attack method targeting multi-label classification models.
In~\cite{kong2}, Kong et al. also proposed a multi-label version of the BA algorithm to address the shortcomings in research on label-only black-box attacks on multi-label models.
Currently, black-box attack methods based on evolutionary algorithms all rely on traditional evolutionary strategies, which suffer from inherent limitations in large-scale optimization, such as high query costs, slow convergence, and difficulty balancing attack performance and perturbation control.

\subsection{Cooperative Co-Evolutionary Algorithms}
The problem of adversarial attacks is inherently a large-scale optimization task, as it involves searching for optimal adversarial perturbations in a high-dimensional feature space.
Cooperative co-evolution provides an efficient solution paradigm by decomposing complex high-dimensional problems into manageable low-dimensional subproblems, making it well-suited for addressing the optimization difficulties in adversarial attack generation.

The early Cooperative Co-evolutionary Algorithm (CCEA) was proposed by Potter and De Jong in 1994 \cite{potter1994cooperative}, which decomposes high-dimensional optimization problems into low-dimensional subproblems solved by independent subpopulations.
Subsequently, Omidvar et al. \cite{omidvar2013cooperative} proposed differential grouping, a data-driven decomposition strategy that automatically discovers interdependent variables and forms near-optimal subcomponents without prior problem knowledge.
This method effectively addresses the curse of dimensionality by revealing the underlying interaction structure of decision variables, significantly improving the performance of evolutionary algorithms on large-scale global optimization problems.
Benefiting from its divide-and-conquer and collaboration mechanisms, numerous CCEA variants have been developed and applied to complex optimization tasks (e.g., large-scale optimization, multi-objective optimization).
Recent advances further enhance CCEA’s performance: Zhou et al. \cite{zhou2023multi} proposed a Multi-Strategy Competitive CCEA (MSCOEA) for multi-objective optimization, balancing uniformity and convergence; Wang et al. \cite{wang2025edge} designed an edge-cloud differential CCEA for distributed high-dimensional feature selection.
Latest studies focus on overcoming key limitations: Qiu et al. \cite{qiu2025novel} proposed a two-phase CCEA framework to handle large-scale optimization with complex overlapping variables, while Guo et al.~\cite{guo2025advancing} introduced a reinforcement learning-guided CCEA to dynamically optimize decomposition strategies.

Nevertheless, the core challenges of CCEA lie in effective variable reduction and rational decomposition strategies. Moreover, CCEA has not been applied to adversarial attack generation, and no related work exists. From the perspective of adversarial attack optimization, it is critical to fully exploit problem characteristics to achieve efficient variable reduction and decomposition for high-dimensional perturbation search. This research gap motivates us to explore CCEA for solving high-dimensional optimization challenges in multi-label label-only black-box attacks.

\section{The MLA-CoE Approach}
\label{sec3}
In the following section, we outline the overall methodology of MLA-CoE for conducting multi-label targeted attacks under the label-only black-box attack scenario.
MLA-CoE generates multi-label black-box adversarial examples through a co-evolutionary algorithm, aiming to effectively attack multi-label classification models in label-only black-box settings by precisely controlling the magnitude of perturbations and reducing query counts. Specifically, MLA-CoE employs a co-evolutionary approach to decompose the complex optimization problem across the entire image space into multiple simpler sub-problems for perturbation optimization. The entire MLA-CoE method is detailed in Algorithm~\ref{alg:MLA-CoE}.
First, we model the optimization problem under the label-only black-box attack setting.
Second, we briefly elaborate on the generation of initial adversarial examples through YOLO initialization.
Next, we introduce the core components of the algorithm: the sparsification process, the co-evolutionary strategy, and the modified DE applied to optimize these sub-problems.

In the label-only black-box attack scenario, attackers face severe challenges due to limited access to model information.
Specifically, the attacker can only observe the output label and cannot obtain confidence scores or the internal mechanism of the model.
Note that this paper focuses on targeted attacks under the label-only black-box setting, and this setting will not be repeated in the following content.

In targeted label-only black-box attacks, such constraints require the attacker to adopt dedicated attack strategies.
The attacker optimizes the perturbation based on a known adversarial example to achieve the expected adversarial goal by only manipulating the output label.
In general, this initial adversarial example is only a feasible solution rather than an optimal one.
The ultimate goal of optimization is to enable the final adversarial example to misclassify the input into the target class with minimal distortion.

Assume that there exists a pre-trained multi-label classifier \(F\) and a clean image \(\mathbf{x} \in \mathbb{R}^d\).
The classifier outputs a label vector denoted as \(F(\mathbf{x}) = \{F_1(\mathbf{x}), F_2(\mathbf{x}), \ldots, F_l(\mathbf{x})\}\),
where \(l\) denotes the total number of labels, and each element \(F_i(\mathbf{x}) \in \{1, -1\}\).
\(F_i(\mathbf{x}) = 1\) indicates that the sample \(\mathbf{x}\) is associated with the \(i\)-th label,
while \(F_i(\mathbf{x}) = -1\) means the sample does not contain this label.
Assume that \(\mathbf{x}^\prime\) is a known adversarial example.
We can model the multi-attack as the following optimization problem.
\begin{equation}
	\label{eq_label_only_attack}
	\begin{array}{ll}
		\mathop{\min} & \|\mathbf{x}^\prime - \mathbf{x}\|_{p}, \\
		\text{s.t.} & F(\mathbf{x}^\prime) = \mathbf{y}^{\star},
	\end{array}
\end{equation}
where \(\|\cdot\|_{p}\) is the \(p\)-norm of the perturbation (typically \(p=2\) or \(p=\infty\) to control the perturbation magnitude),
and \(\mathbf{y}^{\star}\) denotes the desired target label vector.

A necessary condition for solving this optimization problem is to start from a known initial adversarial example.
This initial adversarial example only needs to satisfy the attack objective, even if it differs significantly from the original input.
The ultimate goal of optimization is to continuously reduce the perturbation magnitude so that the adversarial example \(\mathbf{x}^\prime\) gradually approaches the original sample \(\mathbf{x}\).
Meanwhile, the optimization process must ensure that the output of \(F(\mathbf{x}^\prime)\) always meets the constraints of the attack target.
Therefore, generating an effective initial adversarial example is a critical prerequisite to enabling the attack.
To address this challenge effectively, we improve the YOLO initialization proposed by Kong et al.~\cite{kong2} and apply it to generate initial adversarial examples.
Our YOLO initialization method will be detailed in the next section.

\begin{algorithm}[htbp]
	\caption{MLA-CoE Algorithm}
	\label{alg:MLA-CoE}
	\renewcommand{\algorithmicrequire}{\textbf{Input:}}
	\renewcommand{\algorithmicensure}{\textbf{Output:}}
	
	\textbf{Input:} Classifier $F$, input sample $\mathbf{x}$, attack target $\mathbf{y}^\star$, max queries $Q_{\text{max}}$, initial groups $N$, group increment $N_{\delta}$, sparsity probability $q$, sparsification groups $K$, population size $M$, mutation factor $F_{\text{DE}}$, mask init probability $p$, max iterations $T$.\\
	\textbf{Output:} Optimized adversarial example $\mathbf{x}'$
	
	\begin{algorithmic}[1]
		\STATE $\mathbf{x}' = \text{YOLO\_Initialization}(\mathbf{x}, \mathbf{y}^\star)$
		\IF{$F(\mathbf{x}') \neq \mathbf{y}^\star$}
		\STATE \textbf{return} $\mathbf{x}$
		\ELSE
		\STATE $\mathbf{r} = \mathbf{x}' - \mathbf{x}$
		\STATE $\mathbf{r}^{\text{sparse}} = \text{Sparsification}(\mathbf{r}, F, \mathbf{x}, \mathbf{y}^\star, K, q)$ \COMMENT{Alg.~\ref{alg:sparsification}}
		\STATE Initialize query counter $n_q = 0$
		\WHILE{$n_q < Q_{\text{max}}$}
		\STATE $\{\mathcal{G}_1, \dots, \mathcal{G}_N\} = \text{Group}(\mathbf{r}^{\text{sparse}}, N)$
		\FOR{$k = 1$ \textbf{to} $N$}
		\STATE $\mathbf{r}^{\text{sparse}} = \text{DE\_Optimize}(\mathbf{r}^{\text{sparse}}, \mathcal{G}_k, M, F_{\text{DE}}, \mathbf{x}, \mathbf{y}^\star, F, p)$ \COMMENT{Alg.~\ref{alg:de_subproblem}}
		\STATE $n_q = n_q + M \times (T +1)$
		\IF{$n_q > Q_{\text{max}}$}
		\STATE \textbf{break}
		\ENDIF
		\ENDFOR
		\STATE $N = N + N_{\delta}$
		\ENDWHILE
		\STATE $\mathbf{x}' = \mathbf{x} + \mathbf{r}^{\text{sparse}}$
		\STATE \textbf{return} $\mathbf{x}'$
		\ENDIF
	\end{algorithmic}
\end{algorithm}

\subsection{Generating Initial Adversarial Examples by YOLO}
\label{sec:yolo_init}
As mentioned earlier, obtaining valid initial adversarial examples is a fundamental step to solve the constrained optimization problem of label-only black-box attacks.
In this paper, we improve the YOLO initialization method proposed by Kong et al.~\cite{kong2} to generate initial adversarial examples suitable for MLA-CoE.
The detailed steps of the improved YOLO initialization method are as follows.

\textbf{(1) Training the Object Detector:}
In the label-only black-box setting where the training data of the target model is inaccessible, we train a YOLOv3 object detector using a surrogate dataset that shares the same label space $l$ as the target model.
The trained detector is used to locate the relevant regions where target objects are located.
Ultimately, the detector outputs bounding boxes for each detected object, denoted as $B = \{b_1, b_2, \dots, b_k\}$, where $k$ is the number of detected objects and each bounding box $b_i$ represents the coordinates of an object region in the image.

\textbf{(2) Generating Adversarial Candidates:}
We first use the pre-trained YOLOv3~\cite{yolov3} detector to identify all label-related object regions in the original image $\mathbf{x}$, yielding the bounding box set $B = \{b_1, b_2, \dots, b_k\}$.
For each bounding box $b_i$, we map it to the corresponding label (i.e., the label of the object within $b_i$).
Subsequently, we apply Gaussian blur to the regions corresponding to the labels to be removed (as specified in the target label vector $\mathbf{y}^\star$), while keeping the regions corresponding to the labels to be preserved unchanged.
This process yields a raw adversarial candidate $\mathbf{x}_{adv}^{raw}$.

\textbf{(3) Validation with the Target Model:}
The raw adversarial candidate $\mathbf{x}_{adv}^{raw}$ generated in Step 2 is fed into the target model $F$ for validation.
We check whether the output label vector $F(\mathbf{x}_{adv}^{raw})$ matches the target label vector $\mathbf{y}^\star$, i.e., $F(\mathbf{x}_{adv}^{raw}) = \mathbf{y}^\star$.
If this condition is satisfied, a valid candidate adversarial example is successfully generated as $\mathbf{x}_{adv} = \mathbf{x}_{adv}^{raw}$, and we proceed to Step 6 to optimize the initial adversarial example.
If the condition is not met and \textit{labels other than the target ones are hidden} (i.e., some labels that should be preserved are incorrectly removed), we activate the fallback mechanism in Step 4 to repair the adversarial candidate.
If \textit{no labels are hidden at all} (i.e., all original labels are still present), we proceed to Step 5 (Patch Insertion) to generate a new candidate.

\textbf{(4) Fallback Mechanism:}
If the output of the target model $F$ on the blurred image $\mathbf{x}_{adv}^{raw}$ indicates that unexpected labels are hidden, we consider potential causes such as target overlap or ineffective object detection by the YOLOv3 detector.
These scenarios may lead to accidental blurring of regions that should retain their corresponding labels.

For these incorrectly blurred regions, the fallback mechanism repairs them by restoring the original pixel values of these regions from $\mathbf{x}$, yielding a repaired sample $\mathbf{x}_{adv}^{rev}$.
We then input $\mathbf{x}_{adv}^{rev}$ into the target model $F$ again for validation.
If $F(\mathbf{x}_{adv}^{rev}) = \mathbf{y}^\star$, we set $\mathbf{x}_{adv} = \mathbf{x}_{adv}^{rev}$ and proceed to Step 6.
Otherwise (i.e., the fallback mechanism fails to produce a valid adversarial example), we proceed to Step 5 (Patch Insertion).

\textbf{(5) Patch Insertion:}
If both Step 2 and Step 4 fail to generate a valid adversarial candidate, we adopt a patch-based insertion strategy.
First, we apply Gaussian blur to the entire image $\mathbf{x}$, resulting in a fully blurred image where all labels are hidden.
Then, we search for image patches that contain the target labels (i.e., patches that, when inserted into the blurred image, would cause the model to predict the desired labels in $\mathbf{y}^\star$).
These patches are pasted into arbitrary positions of the blurred image to generate a new candidate $\mathbf{x}_{adv}^{patch}$.
We then input $\mathbf{x}_{adv}^{patch}$ into the target model $F$ for validation.
If $F(\mathbf{x}_{adv}^{patch}) = \mathbf{y}^\star$, we set $\mathbf{x}_{adv} = \mathbf{x}_{adv}^{patch}$ and proceed to Step 6.
Otherwise, the YOLO initialization is deemed to have failed.

\textbf{(6) Binary Search Optimization:}
Once a valid initial adversarial example $\mathbf{x}_{adv}$ is obtained, we perform a binary search between $\mathbf{x}_{adv}$ and the original image $\mathbf{x}$ to optimize the perturbation.

The mathematical formulation of the binary search process is given as follows:
\begin{equation}
	\mathbf{x}^\prime = \mathbf{x} + \alpha \cdot \frac{\mathbf{x}_{adv} - \mathbf{x}}{\|\mathbf{x}_{adv} - \mathbf{x}\|_2}, \quad \alpha \in [0, \|\mathbf{x}_{adv} - \mathbf{x}\|_2],
\end{equation}
where $\alpha$ is the step size, $\mathbf{x}^\prime$ is the final initial adversarial example, $\mathbf{x}$ is the original input.
The binary search operates on the scalar step size $\alpha$ by initializing the search interval with $\alpha_{low} = 0$ and $\alpha_{high} = \|\mathbf{r}\|_2$, then iteratively computing the midpoint $\alpha_{mid} = (\alpha_{low} + \alpha_{high}) / 2$ and updating the bounds based on the adversarial check: if $\mathbf{x}^\prime(\alpha_{mid})$ successfully fools the model, the upper bound is reduced to $\alpha_{high} = \alpha_{mid} - \epsilon$ to seek a smaller perturbation; otherwise, the lower bound is increased to $\alpha_{low} = \alpha_{mid} + \epsilon$ to strengthen the attack, where $\epsilon$ denotes the convergence threshold (e.g., $10^{-5}$) controlling the precision of the binary search.

The process continues until the maximum allowed number of queries is reached or the search interval collapses, after which the adversarial example with the smallest successful $\alpha$ is used as the initial input for subsequent MLA-CoE operations.

It is worth noting that for samples where initialization fails in the above process, we adopt the original YOLO initialization method proposed by Kong et al.~\cite{kong2} as an alternative to re-attempt initialization.

\subsection{Sparsification Strategy}
\label{sec:sparsification}
To further reduce the optimization space and enhance efficiency, we introduce a sparsification strategy before the cooperative co-evolution process, as detailed in Algorithm~\ref{alg:sparsification}.
The initial adversarial example usually differs significantly from the original image $\mathbf{x}$ only in local regions, which means the perturbation is not distributed across the entire image.
In addition, the initial adversarial example is merely a feasible solution to the optimization problem rather than the unique solution. 
Under the assumption that an adversarial example with imperceptible perturbations exists for each sample, the optimal perturbation is expected to be more sparse than that of the initial adversarial example.  
Therefore, prior to the commencement of cooperative co-evolution, we apply sparsification to the initial perturbation to commence optimization from a minimal search space, thereby significantly reducing the problem complexity.
\begin{algorithm}[htbp]
	\caption{Sparsification Strategy}
	\label{alg:sparsification}
	\renewcommand{\algorithmicrequire}{\textbf{Input:}}
	\renewcommand{\algorithmicensure}{\textbf{Output:}}
	
	\textbf{Input:} Initial perturbation $\mathbf{r}$, original image $\mathbf{x}$, classifier $F$, target $\mathbf{y}^\star$, number of groups $K$, sparsity probability $q$.\\
	\textbf{Output:} Sparse perturbation $\mathbf{r}^{\text{sparse}}$
	
	\begin{algorithmic}[1]
		\STATE $\mathcal{I} = \{i \mid r_i \neq 0\}$
		\STATE $\{\mathcal{I}_1, \dots, \mathcal{I}_K\} \leftarrow \text{Partition}(\mathcal{I}, K)$
		\STATE $\mathbf{m} = \mathbf{1} ^d$
		
		\FOR{$k = 1$ \textbf{to} $K$}
		\STATE $\tilde{\mathbf{m}} = \mathbf{m}$
		\FOR{each $i \in \mathcal{I}_k$}
		\STATE $\xi \sim \mathcal{U}(0, 1)$
		\IF{$\xi < q$}
		\STATE $\tilde{m}_i = 0$
		\ENDIF
		\ENDFOR
		\STATE $\tilde{\mathbf{r}} = \tilde{\mathbf{m}} \odot \mathbf{r}$
		\STATE $\tilde{\mathbf{x}} = \mathbf{x} + \tilde{\mathbf{r}}$
		\IF{$F(\tilde{\mathbf{x}}) = \mathbf{y}^\star$}
		\STATE $\mathbf{m} = \tilde{\mathbf{m}}$
		\ENDIF
		\ENDFOR
		
		\STATE $\mathbf{r}^{\text{sparse}} = \mathbf{m} \odot \mathbf{r}$
		\STATE \textbf{return} $\mathbf{r}^{\text{sparse}}$
	\end{algorithmic}
\end{algorithm}

Specifically, given the initial perturbation $\mathbf{r}$, we first identify the set of non-zero indices $\mathcal{I} = \{i \mid r_i \neq 0\}$ (Line 1 of Algorithm~\ref{alg:sparsification}). 
These indices are divided into $K$ disjoint groups $\{\mathcal{I}_1, \mathcal{I}_2, \dots, \mathcal{I}_K\}$ (Line 2) with approximately equal sizes. Specifically, the first $K-1$ groups each contain $\lfloor |\mathcal{I}|/K \rfloor$ elements, while the last group contains $\lfloor |\mathcal{I}|/K \rfloor + (|\mathcal{I}| \bmod K)$ elements.
We initialize a binary mask vector $\mathbf{m}$ with all elements set to $1$ (Line 3). The sparsification process proceeds by traversing each group sequentially (Lines 4--17). For group $\mathcal{I}_k$, we create a temporary mask $\tilde{\mathbf{m}}$ by copying the current mask $\mathbf{m}$ (Line 5). For each index $i \in \mathcal{I}_k$, we set $\tilde{m}_i = 0$ with probability $q$ (Lines 6--11) to progressively remove redundant perturbations and encourage sparsity.

The candidate sparse perturbation is computed as $\tilde{\mathbf{r}} = \tilde{\mathbf{m}} \odot \mathbf{r}$ (Line 12), where $\odot$ denotes element-wise multiplication. We then construct the candidate adversarial example $\tilde{\mathbf{x}} = \mathbf{x} + \tilde{\mathbf{r}}$ (Line 13) and query the target model $F$ to evaluate whether the attack goal is achieved (i.e., $F(\tilde{\mathbf{x}}) = \mathbf{y}^\star$). 
If the sparsified perturbation $\tilde{\mathbf{r}}$ preserves adversariality, we accept the sparsification by updating the mask $\mathbf{m} \leftarrow \tilde{\mathbf{m}}$ (Lines 14--16); otherwise, we retain the original mask and discard the modifications to group $\mathcal{I}_k$. After processing all $K$ groups, the final sparse perturbation is obtained as $\mathbf{r}^{\text{sparse}} = \mathbf{m} \odot \mathbf{r}$ (Line 18).

\subsection{Co-evolution Strategy}
Based on the positions of these non-zero entries, we first perform the grouping operation on the sparse perturbation $\mathbf{r}^{\text{sparse}}$, and the grouping process can be formally expressed as.
\begin{equation}
	\label{eq_grouping}
	\mathbf{r}_i = \mathbf{r}^{\text{sparse}} \odot \mathbb{I}(\mathcal{G}_i), \quad i = 1, 2, \dots, N
\end{equation}
where $\mathbf{r}_i$ denotes the $i$-th subcomponent of the perturbation after grouping, $\odot$ represents the element-wise multiplication operation, $\mathbb{I}(\mathcal{G}_i)$ is the indicator function (i.e., $\mathbb{I}(\mathcal{G}_i)(p) = 1$ if the position $p$ of $\mathbf{r}^{\text{sparse}}$ belongs to the $i$-th group $\mathcal{G}_i$, otherwise $\mathbb{I}(\mathcal{G}_i)(p) = 0$), and $N$ is the total number of groups for co-evolution.

After grouping, we employ DE to optimize the pixels within each group independently, as described in Algorithm~\ref{alg:de_subproblem}.
In the DE optimization process, population initialization and mutation operations are executed within each group, and the crossover operation is not adopted.

Notably, the evaluation and selection operations are not performed independently.
We evaluate the complete perturbation formed by the cooperative collaboration of all subcomponents, and conduct the selection process based on the evaluation results of the global perturbation.
The detailed cooperative assembly strategy is formulated as follows.
\begin{equation}
	\label{eq_segmentation}
	\mathbf{r}^{\text{sparse}} = \mathbf{r}_1 \oplus \mathbf{r}_2 \oplus \cdots \oplus \mathbf{r}_N,
\end{equation}
where $\oplus$ denotes the concatenation (or assembly) operation that reconstructs the full perturbation from subcomponents; $\mathbf{r}^{\text{sparse}}$ here represents the current global perturbation being optimized; $\mathbf{r}_1, \mathbf{r}_2, \dots, \mathbf{r}_N$ are the $N$ disjoint subcomponents obtained after grouping.

Within each group, population initialization and mutation operations are performed locally and independently, with no crossover operation.
The fitness of each individual in a subpopulation is assessed using the \emph{global perturbation} formed by cooperatively assembling all subcomponents.
After all groups have completed one iteration, the number of groups $N$ is increased by $N_{\delta}$.
This divide-and-conquer and collaborative evaluation mechanism enables efficient large-scale optimization while ensuring the final perturbation remains globally effective.

\subsection{Differential Evolution for Subproblem Optimization}
\label{MLA-CoE}
Our MLA-CoE adopts DE to optimize each group of indices, as presented in Algorithm~\ref{alg:de_subproblem}. To facilitate the cooperative strategy, DE operates on \textbf{binary masks} rather than directly on perturbation values. Let $\mathbf{r}^{\text{sparse}}$ denote the sparsified perturbation vector obtained from Algorithm~\ref{alg:sparsification}, and let $\mathcal{G}$ be the set of indices of its non-zero elements. 
These indices are divided into $N$ groups $\{\mathcal{G}_1, \mathcal{G}_2, \dots, \mathcal{G}_N\}$ for co-evolution. 
For each group $\mathcal{G}_k$, we maintain a population of masks $\mathbf{pop}_k = \{\mathbf{p}_1, \mathbf{p}_2, \dots, \mathbf{p}_M\}$, where each individual is initialized as $\mathbf{p}_j \in \{0,1\}^d$ a binary vector, constructed as follows (Line 1 of Algorithm~\ref{alg:de_subproblem}).
\begin{equation}
	\mathbf{p}_j[i] = 
	\begin{cases} 
		0, & \text{if } i \in \mathcal{G}_k \text{ and } \xi < p, \\
		1, & \text{Otherwise},
	\end{cases}
\end{equation}
where $\mathbf{p}_j[i]$ is the $i$-th component of the $j$-th individual, $\xi \sim \mathcal{U}(0,1)$. 
That is, for indices within the target group $\mathcal{G}_k$, the mask value is set to $0$ with probability $p$ and $1$ with probability $1-p$; for all other groups, the mask values are fixed to $1$. 
Consequently, each individual $\mathbf{p}_j$ serves as a complete mask that can be directly applied to $\mathbf{r}^{\text{sparse}}$ via element-wise multiplication $\mathbf{p}_j \odot \mathbf{r}^{\text{sparse}}$ to produce the final perturbation for evaluation.
\begin{algorithm}[htbp]
	\caption{Modified Differential Evolution}
	\label{alg:de_subproblem}
	\renewcommand{\algorithmicrequire}{\textbf{Input:}}
	\renewcommand{\algorithmicensure}{\textbf{Output:}}
	
	\textbf{Input:} Current group number $k$, current perturbation $\mathbf{r}^{\text{sparse}}$, index set of group $k$ denoted as $\mathcal{G}_k$, population size $M$, differential weight $F_{\text{DE}}$, maximum number of generations $T$, original image $\mathbf{x}$, target label $\mathbf{y}^\star$, target model $F$, and probability $p$.\\
	\textbf{Output:} Updated $\mathbf{r}^{\text{sparse}}$
	
	\begin{algorithmic}[1]
		\STATE $\mathbf{pop}_k = \{\mathbf{p}_j\}_{j=1}^M$ where $\mathbf{p}_j \in \mathbb{R}^{|\mathcal{G}_k|}$
		\STATE Evaluate $\text{fitness}(\mathbf{pop}_k)$ via Eq. \eqref{fitness}, $\forall j$
		\STATE $t = 0$
		\WHILE{$t < T$}
		\FOR{$j = 1$ \textbf{to} $M$}
		\STATE Select $\mathbf{p}_a, \mathbf{p}_b \in \mathbf{pop}_k$ ($a \neq b \neq j$)
		\STATE $\mathbf{off}_j = \mathbf{p}_j + F_{\text{DE}} \cdot (\mathbf{p}_a - \mathbf{p}_b)$ \COMMENT{Eq. \eqref{eq_mutation}}
		 
		\ENDFOR
		\STATE Evaluate $\text{fitness}(\mathbf{off})$ via Eq. \eqref{fitness}
		\FOR{$j = 1$ \textbf{to} $M$}
		\STATE $\mathbf{p}_j \leftarrow \begin{cases} \mathbf{off}_j, & \text{if } \text{fitness}(\mathbf{off}_j) < \text{fitness}(\mathbf{p}_j) \\ \mathbf{p}_j, & \text{otherwise} \end{cases}$ \COMMENT{Eq. \eqref{eq_selection}}
		\ENDFOR
		\STATE $t \leftarrow t + 1$
		\ENDWHILE
		\STATE $\mathbf{p}_{\text{best}} = \arg\min_{\mathbf{p} \in \mathbf{pop}_k} \text{fitness}(\mathbf{p})$
		\STATE $\mathbf{r}^{\text{sparse}}[\mathcal{G}_k] \leftarrow \mathbf{p}_{\text{best}} \odot \mathbf{r}^{\text{sparse}}[\mathcal{G}_k]$ \COMMENT{Update group $\mathcal{G}_k$ only}
		\STATE \textbf{return} $\mathbf{r}^{\text{sparse}}$
	\end{algorithmic}
\end{algorithm}

\textbf{Fitness Evaluation:}
The fitness function evaluates the quality of each individual $\mathbf{p}_j$ (candidate solution for group $\mathcal{G}_k$) by measuring the perturbation magnitude of the resulting adversarial example. 
Specifically, $\mathbf{p}_j$ is applied to the $k$-th group of $\mathbf{r}^{\text{sparse}}$ (i.e., $\mathbf{r}_k$) to construct the adversarial example $\mathbf{x}' = \mathbf{x} + \mathbf{r}^{\text{sparse}}$. The fitness evaluation function is defined as follows:
\begin{equation}
	\label{fitness}
	\text{fitness}(\mathbf{p}_j) =
	\begin{cases}
		\|\mathbf{r}^{\text{sparse}}\|_2  & \text{if } F(\mathbf{x}') = \mathbf{y}^{\star}, \\
		+\infty & \text{Otherwise}.
	\end{cases}
\end{equation}
The significance of the fitness evaluation function is that if the attack goal is achieved, the fitness equals the $L_2$ norm of the perturbation; otherwise, the individual receives infinite fitness and is discarded. 
Therefore, in this paper, smaller fitness implies a smaller $L_2$ norm of the perturbation vector, ultimately driving the evolution toward minimal perturbations.

\textbf{Mutation Operator:}
We employ the DE/rand/1 mutation strategy to generate offspring (Lines 5--8 of Algorithm~\ref{alg:de_subproblem}). For each individual $\mathbf{p}_j$ in population $\mathbf{pop}_k$, the mutated offspring $\mathbf{off}_j$ is generated as:
\begin{equation}
	\label{eq_mutation}
	\mathbf{off}_j = \mathbf{p}_j + F_{\text{DE}} \cdot (\mathbf{p}_a - \mathbf{p}_b),
\end{equation}
where $F_{\text{DE}}$ is the mutation factor, which is set to 0.4 in this paper, and $\mathbf{p}_a, \mathbf{p}_b$ are two distinct individuals randomly selected from the current population ($a, b \neq j$ and $a \neq b$). Notably, since individuals in $\mathbf{pop}_k$ already have fixed values of $1$ outside $\mathcal{G}_k$, mutation naturally preserves these regions while only modifying positions within $\mathcal{G}_k$.

\textbf{Selection Operator:}
Following the survival-of-the-fittest principle, the selection operator chooses the better individual between each parent and its corresponding offspring to retain in the next generation population (Line 11 of Algorithm~\ref{alg:de_subproblem}). For each pair $(\mathbf{p}_j, \mathbf{off}_j)$, the selection operator is defined as follows:
\begin{equation}
	\label{eq_selection}
	\mathbf{p}_j \leftarrow 
	\begin{cases}
		\mathbf{off}_j & \text{if } \text{fitness}(\mathbf{off}_j) < \text{fitness}(\mathbf{p}_j), \\
		\mathbf{p}_j & \text{Otherwise}.
	\end{cases}
\end{equation}
After updating population $\mathbf{pop}_k$, the best individual $\mathbf{p}_{\text{best}} = \arg\min_{\mathbf{p} \in \mathbf{pop}_k} \text{fitness}(\mathbf{p})$ (Line 15 of Algorithm~\ref{alg:de_subproblem}) is used to update the global perturbation: $\mathbf{r}^{\text{sparse}} \leftarrow \mathbf{p}_{\text{best}} \odot \mathbf{r}^{\text{sparse}}$ (Line 16 of Algorithm~\ref{alg:de_subproblem}), ensuring subsequent groups optimize based on the improved solution.

\textbf{Termination:} 
The algorithm terminates when the maximum number of iterations $T$ is reached or the query budget is exhausted.

\section{Experimental Setup}
\label{sec4}

This section outlines the structure of the experiments, detailing the datasets, models, and parameter configurations utilized. We will begin by describing the datasets, emphasizing their characteristics such as sample size, label distribution, and data origin. Following this, we will present the models employed and their specific parameter settings.

The source codes of this paper are available at https://anonymous.4open.science/r/MLA-CoE/.
\subsection{Datasets}
For our experiments, we adopt two widely-used multi-label image benchmark datasets: PASCAL VOC 2007 \cite{VOC2007} and PASCAL VOC 2012 \cite{VOC2012}.

\textbf{PASCAL VOC 2007}: As a classic benchmark for multi-label image classification, it contains 5,011 training images, 2,510 validation images, and 4,952 test images, covering 20 object categories.

\textbf{PASCAL VOC 2012}: As an extended version of VOC 2007, it follows the same 20 label categories and contains 5,017 training samples and 5,821 validation samples, without an official test set.

The detailed statistics of these two datasets are summarized in Table \ref{tab:data}.
Since VOC 2012 does not provide an official test split, we randomly divide its original training set into 80\% for training and 20\% for validation, and use its original validation set as the test set in our experiments.
All images are uniformly resized to 448×448×3, leading to a high-dimensional optimization space of 602,112 dimensions, which brings great challenges to differential evolution for multi-label black-box adversarial attacks.
\begin{table}[htbp]
	\centering
	\caption{Dataset Statistics}
	\resizebox{\columnwidth}{!}{
		\begin{tabular}{c|c|c|c}
			\hline
			Dataset & Training and Validation Samples & Testing Samples & \# Labels \\
			\hline
			VOC 2007 & 7,521 & 4,952 & 20 \\
			VOC 2012 & 5,017 & 5,821 & 20 \\
			\hline
		\end{tabular}%
	}
	\label{tab:data}%
\end{table}%

\subsection{Models}
We conduct all experiments on two representative multi-label classification models: ML-LIW \cite{song} and ML-GCN \cite{ML-GCN}.

\textbf{ML-LIW}: Proposed by Song et al. \cite{song}, ML-LIW is built upon the Inception-v3 backbone \cite{inceptionv3} pre-trained on ImageNet \cite{imagenet}.
To adapt to multi-label classification, it replaces the softmax layer with a sigmoid layer and retrains the classification module.
The model integrates instance-level loss and label-level loss to maintain stable and reliable classification performance.

\textbf{ML-GCN}: Introduced by Chen et al. \cite{ML-GCN}, ML-GCN adopts a graph convolutional network (GCN) to explicitly capture label dependencies.
It constructs a directed graph where nodes represent label word embeddings, and uses GCN to encode label correlations, greatly enhancing the representation ability of the model.
The original GCN structure is carefully redesigned to fit multi-label learning tasks.

In our implementation, both models strictly follow the training settings in their original papers.
ML-LIW is trained with the Adam optimizer for 100 epochs at a learning rate of 0.001.
ML-GCN is optimized by SGD for 100 epochs with an initial learning rate of 0.01, which decays by 90\% every 40 epochs.
The classification performance of ML-GCN and ML-LIW on VOC 2007 and VOC 2012 datasets is reported in Table \ref{ACC}.

\begin{table}[htbp]
	\centering
	\caption{Performance of ML-GCN and ML-LIW on Datasets}
	\begin{tabular}{c|c|c}
		\hline
		Models & Datasets & Average Precision \\
		\hline
		\multirow{2}[0]{*}{ML-GCN} & VOC 2007 & 0.9686  \\
		& VOC 2012 & 0.9586  \\
		\hline
		\multirow{2}[0]{*}{ML-LIW} & VOC 2007 & 0.9031  \\
		& VOC 2012 & 0.9356  \\

		\hline
	\end{tabular}%
	\label{ACC}%
\end{table}%

\subsection{Parameter Settings}
To ensure fair comparison, all methods are limited to a maximum of 10,000 queries. The $L_2$-norm constraint is set to $38.798$ (\( \sqrt{3 \times 448 \times 448} \times 0.05  \approx 38.798\)), meaning an attack is considered successful only when the perturbation is below this threshold. 
Furthermore, all methods adopt a unified YOLO initialization strategy, where the maximum number of queries for YOLO initialization is set to 200. 
The Gaussian blur radius was set to 15.
This configuration ensures the consistency of the initial state across all methods, thereby guaranteeing the fairness of experimental comparisons.
For a fair comparison, the maximum number of queries is uniformly limited to 10,000 for all methods.

\textbf{Parameters:}
\begin{itemize}
	\item \textbf{MLA-CoE:} Population size $M=20$, sparsification probability $q=0.75$ sparsification groups $K=1000$, perturbation groups $N=50$, mask init probability $p=0.5$, mutation factor $F_{\text{DE}}=0.4$. MLA-CoE/$N$ denotes configurations with varying group numbers (e.g., MLA-CoE/25 means $N=25$).
	\item \textbf{ML-LBA~\cite{kong2}:} Population size $=100$. All other parameters remain the default settings in~\cite{kong2}.
	\item \textbf{ML-BA~\cite{kong2}:} All parameters follow the default settings in~\cite{kong2}.
\end{itemize}

\textbf{Attack Type:} We evaluate the ``hiding single" attack. For each sample containing two or more labels, one positive label is randomly selected to be hidden. We select 100 samples from each dataset that meet this criterion for evaluation.

\section{Results and Analysis}
\label{sec5}
This section comprehensively evaluates the performance of three label-only attack methods (MLA-CoE, ML-LBA, ML-BA) on two representative multi-label models (ML-GCN and ML-LIW) across VOC 2007 and VOC 2012 datasets. The evaluation focuses on two core dimensions: \textit{Attack Success Rate (ASR)} (to measure attack effectiveness, denoted as $\uparrow$, indicating the higher the value, the better the performance) and \textit{perturbation norm metrics} (including $L_1$, $L_2$, $L_{\infty}$, RSMD (Root Mean Square Deviation), and perturbation mean, to quantify attack stealthiness, all denoted as $\downarrow$, indicating the lower the value, the better the stealthiness).

\subsection{Comprehensive Evaluation of Adversarial Attack Performance}
\subsubsection{Attack Success Rate (ASR) Analysis}
ASR is the primary metric for assessing the effectiveness of adversarial attacks, representing the percentage of adversarial examples that successfully mislead the target model. 

As shown in Table~\ref{perturbation_results}, MLA-CoE outperforms other methods consistently, achieving the highest ASR across all model-dataset combinations and thus demonstrating superior attack capability in label-only settings. For the ML-GCN model, MLA-CoE achieves an ASR of 81\% on the VOC 2007 dataset and 91\% on the VOC 2012 dataset; for the ML-LIW model, its ASR reaches 66\% on VOC 2007 and 73\% on VOC 2012. 

ML-LBA exhibits moderate effectiveness, with intermediate ASR values ranging from 30\% for the ML-LIW model on VOC 2012 to 51\% for the ML-GCN model on VOC 2012. ML-LBA outperforms ML-BA across all model-dataset combinations, with notable differences including an ASR of 41\% for ML-LBA compared to 22\% for ML-BA on the ML-GCN model with the VOC 2007 dataset, and 33\% for ML-LBA compared to 14\% for ML-BA on the ML-LIW model with the VOC 2007 dataset.

ML-BA is the least effective, achieving the lowest ASR in all experimental scenarios. Specifically, its ASR values are 22\% on the ML-GCN model with VOC 2007, 35\% on the ML-GCN model with VOC 2012, 14\% on the ML-LIW model with VOC 2007, and 21\% on the ML-LIW model with VOC 2012. ML-BA never outperforms ML-LBA in any model-dataset combination and remains far behind MLA-CoE.
\begin{table*}[ht]
	\centering
	\caption{Performance Comparison of MLA-CoE with State-of-the-art methods (ML-LAB, ML-BA)}
	\resizebox{\linewidth}{!}{
		\begin{tabular}{c|c|c|c|c|c|c|c|c}
			\hline
			\textbf{Model} & \textbf{Datasets} & \textbf{Method} & \textbf{ASR $\uparrow$} & \textbf{$L_1$ Norm $\downarrow$} & \textbf{$L_2$ Norm $\downarrow$} & \textbf{$L_\infty$ Norm $\downarrow$} & \textbf{RSMD $\downarrow$} & \textbf{Mean $\downarrow$} \\
			\hline
			\multirow{6}{*}{ML-GCN} & \multirow{3}{*}{VOC 2007} 
			& MLA-CoE & \textbf{81\%} & \textbf{2143.8741} & \textbf{15.4785} & \textbf{0.5200} & \textbf{2.5433} & \textbf{0.0036} \\
			&                         & ML-LBA & 41\% & 3317.9869 & 19.0044 & 0.5411 & 3.1227 & 0.0055 \\
			&                         & ML-BA  & 22\% & 4873.5558 & 23.6819 & 0.7574 & 3.8913 & 0.0081 \\
			\cline{2-9}
			& \multirow{3}{*}{VOC 2012} 
			& MLA-CoE & \textbf{91\%} & \textbf{1736.1275} & \textbf{13.4997} & \textbf{0.4593} & \textbf{2.2182} & \textbf{0.0029} \\
			&                         & ML-LBA & 51\% & 2956.8044 & 15.3832 & 0.4800 & 2.5277 & 0.0049 \\
			&                         & ML-BA  & 35\% & 3981.0751 & 18.5796 & 0.6697 & 3.0529 & 0.0066 \\
			\hline
			\multirow{6}{*}{ML-LIW} & \multirow{3}{*}{VOC 2007} 
			& MLA-CoE & \textbf{66\%} & \textbf{2764.6858} & \textbf{18.6392} & \textbf{0.5232} & \textbf{3.0627} & \textbf{0.0046} \\
			&                         & ML-LBA & 33\% & 4975.5312 & 21.9796 & 0.5571 & 3.6115 & 0.0083 \\
			&                         & ML-BA  & 14\% & 3742.8953 & 16.3806 & 0.4588 & 2.6915 & 0.0062 \\
			\cline{2-9}
			& \multirow{3}{*}{VOC 2012} 
			& MLA-CoE & \textbf{73\%} & \textbf{2907.0594} & \textbf{18.2506} & \textbf{0.5165} & \textbf{2.9988} & \textbf{0.0048} \\
			&                         & ML-LBA & 30\% & 4042.2614 & 20.8984 & 0.5513 & 3.4339 & 0.0067 \\
			&                         & ML-BA  & 21\% & 4867.8754 & 20.6373 & 0.6528 & 3.3910 & 0.0081 \\
			\hline
		\end{tabular}
	}
	\label{perturbation_results}
\end{table*}

\subsubsection{Perturbation Analysis}
Perturbation norms quantify the magnitude of modifications to original images, with smaller values indicating better stealthiness. 

As presented in Table~\ref{perturbation_results}, MLA-CoE maintains minimal perturbation, achieving the lowest perturbation across all norm metrics in nearly all experiments (except for the ML-LIW model on VOC 2007, where ML-BA shows slightly lower $L_2$ and $L_\infty$ norms). This exception is attributable to the extremely low ASR of ML-BA (14\%), meaning that only a small fraction of successfully attacked samples (14 out of 100) contributed to the average perturbation calculation, making the metric statistically unreliable and unrepresentative of the method's actual performance. Excluding this case, MLA-CoE demonstrates a favorable trade-off between attack effectiveness and stealthiness. For the ML-GCN model on the VOC 2012 dataset, the $L_2$ norm of MLA-CoE is 12.2\% lower than that of ML-LBA and 27.3\% lower than that of ML-BA; for the ML-LIW model on the VOC 2007 dataset, the $L_1$ norm of MLA-CoE is only 73.9\% of that of ML-LBA. Additionally, the $L_{\infty}$ norm and perturbation mean of MLA-CoE are consistently the smallest among all methods, and its RSMD is significantly lower than those of ML-LBA and ML-BA. 

ML-LBA exhibits moderate perturbation levels, with its perturbation norms falling between those of MLA-CoE and ML-BA in most cases. For instance, on the ML-GCN model with the VOC 2007 dataset, the $L_2$ norm of ML-LBA is 22.8\% higher than that of MLA-CoE but 19.8\% lower than that of ML-BA, while its RSMD is 22.8\% higher than that of MLA-CoE but 19.8\% lower than that of ML-BA.

ML-BA incurs excessive perturbation in most scenarios, exhibiting the largest perturbation across most metrics. On the ML-LIW model with the VOC 2007 dataset, the $L_1$ norm of ML-BA is 1.35 times that of MLA-CoE, and its perturbation mean is also 1.35 times higher than that of MLA-CoE; on the ML-LIW model with the VOC 2012 dataset, the $L_2$ norm of ML-BA is 13.1\% higher than that of MLA-CoE. This excessive perturbation renders the adversarial examples generated by ML-BA easily detectable, thereby limiting their practical applicability.

\subsubsection{Visualization and Performance Analysis}
\begin{figure}[t]
	\centering
	
	\small
	\setlength{\tabcolsep}{3pt}
	\renewcommand{\arraystretch}{0.5}
	
	\begin{tabular}{ccc}
		\includegraphics[width=0.15\textwidth]{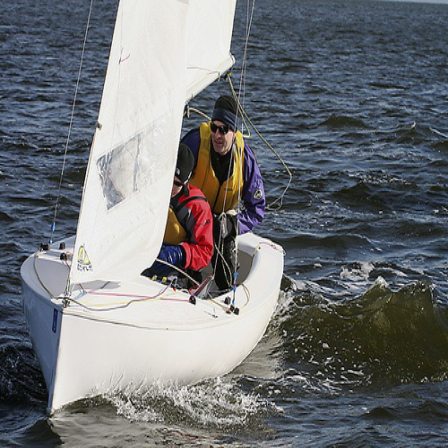} & 
		\includegraphics[width=0.15\textwidth]{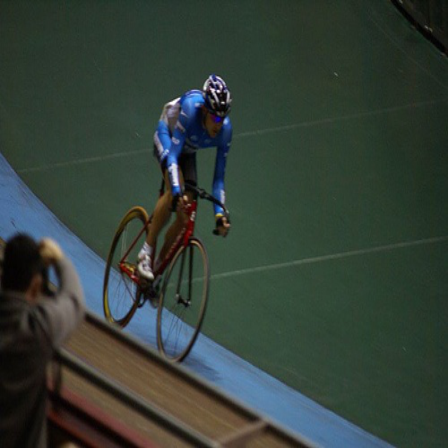} & 
		\includegraphics[width=0.15\textwidth]{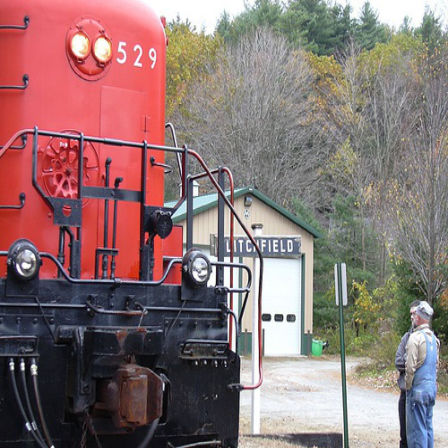} \\
		\scriptsize boat, person & \scriptsize bicycle, person & \scriptsize person, train \\[2pt]
		
		\includegraphics[width=0.15\textwidth]{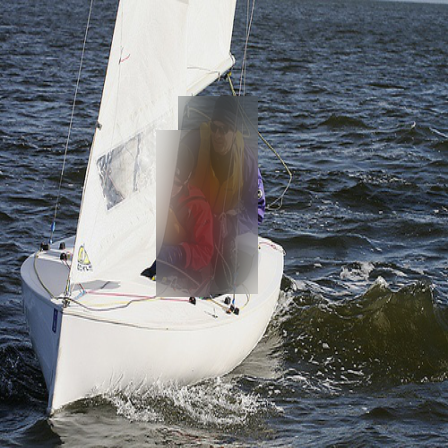} & 
		\includegraphics[width=0.15\textwidth]{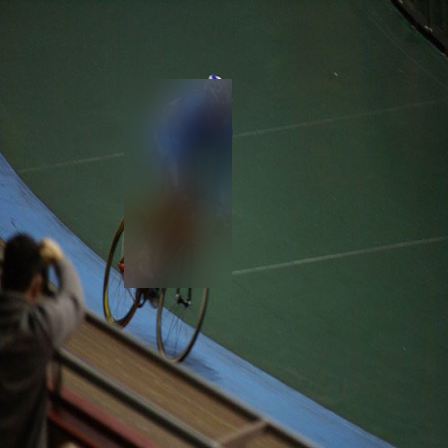} & 
		\includegraphics[width=0.15\textwidth]{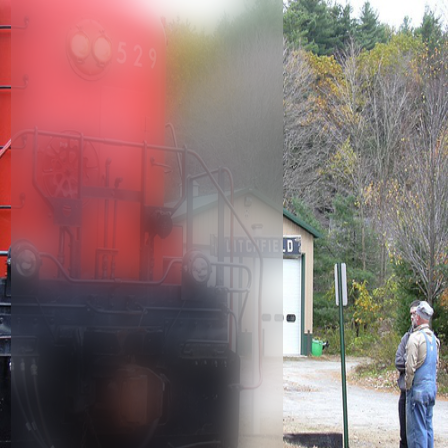} \\
		\scriptsize boat & \scriptsize bicycle & \scriptsize person \\[2pt]
		
		\includegraphics[width=0.15\textwidth]{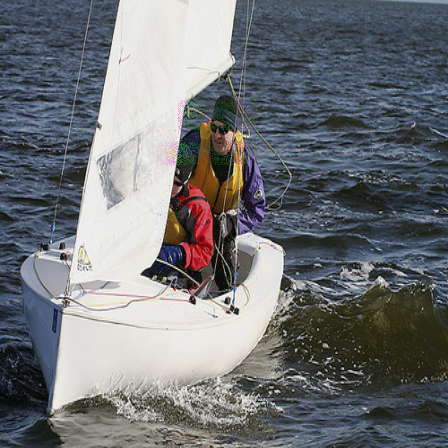} & 
		\includegraphics[width=0.15\textwidth]{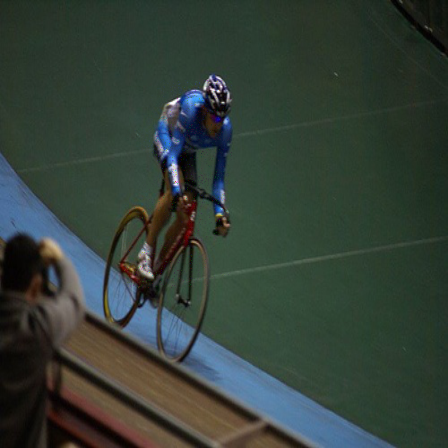} & 
		\includegraphics[width=0.15\textwidth]{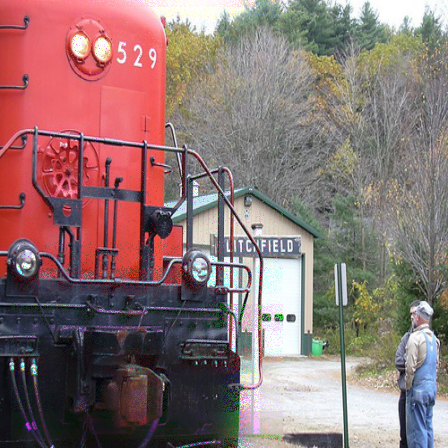} \\
		\scriptsize boat & \scriptsize bicycle & \scriptsize person \\
		\scriptsize $L_2$ = 10.12 & \scriptsize $L_2$ = 6.17 & \scriptsize $L_2$ = 25.15 \\
	\end{tabular}
	
	\vspace{-6pt}
	\caption{\small Generation of multi-label adversarial examples by MLA-CoE. The first row displays original images with their ground-truth labels; the second row shows initial adversarial examples; the third row presents the final adversarial examples generated by MLA-CoE. The predicted labels of the multi-label classifier are shown below each image, and the $L_2$ norm of each adversarial perturbation is indicated beneath the predicted labels for the final examples.}
	\label{fig:mla_coe_adv_examples}
	\vspace{-6pt}
\end{figure}

\begin{figure*}[t]
	\centering
	\includegraphics[width=0.7\textwidth]{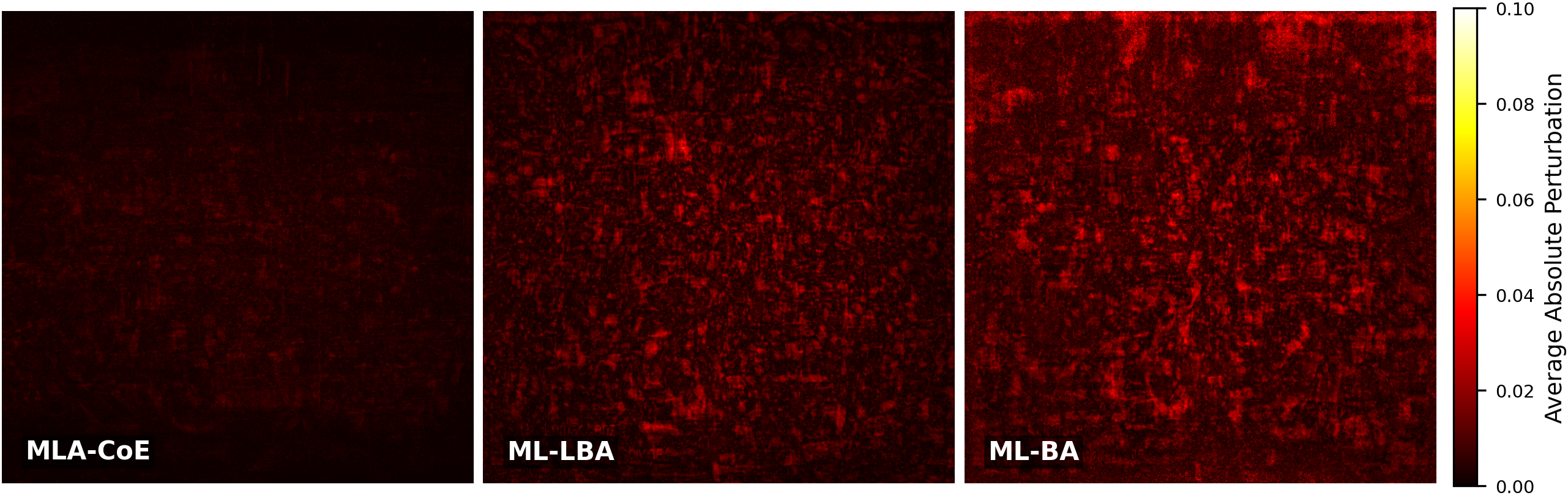} 
	\caption{Pixel-wise average absolute perturbation of adversarial examples generated by three label-only attack methods (MLA-CoE, ML-LBA, and ML-BA) on the ML-GCN model over the VOC 2007 dataset, calculated as the mean perturbation of all successfully attacked samples. Brighter regions indicate larger modifications, darker regions smaller imperceptible perturbations, with a shared colorbar for consistent comparison.}
	\label{fig:perturbation_heatmap}
\end{figure*}

The effectiveness, stealthiness, and comprehensive performance of the proposed MLA-CoE method are validated through both qualitative and quantitative visualization results, as illustrated in Fig.~\ref{fig:mla_coe_adv_examples} and Fig.~\ref{fig:perturbation_heatmap}.

Fig.~\ref{fig:mla_coe_adv_examples} illustrates the process of MLA-CoE generating adversarial examples. As can be seen in the figure, MLA-CoE can generate visually imperceptible multi-label adversarial images while achieving excellent attack effects.

Fig.~\ref{fig:perturbation_heatmap} shows the heatmap of the perturbation mean of the three attack methods on the ML-GCN-VOC 2007 dataset. It can be seen intuitively from the figure that the perturbation mean introduced by MLA-CoE is the smallest.
This result demonstrates that MLA-CoE outperforms ML-LBA and ML-BA in balancing high attack success rates and low perturbation intensity in most scenarios.

\subsection{Analysis of The Impact of Group Number}

To investigate the impact of group number on the performance of MLA-CoE, we conduct ablation experiments with varying group numbers (25, 50, 75, 100, 150, 200) across two models (ML-GCN, ML-LIW) and two datasets (VOC 2007, VOC 2012), with results summarized in Table \ref{tab:ablation_group}.

Overall, the group number exhibits a non-monotonic influence on both attack effectiveness (ASR) and perturbation intensity (e.g., $L_1$/ $L_2$ norm, perturbation mean). For the ML-GCN model, MLA-CoE with 50 groups achieves the highest ASR on both VOC 2007 (81\%) and VOC 2012 (91\%), representing the optimal balance between attack success rate and perturbation cost. When the group number is less than 50 (e.g., 25 groups), ASR is slightly lower (78\% on VOC 2007, 89\% on VOC 2012) despite reduced perturbation metrics (e.g., $L_2$ norm = 14.0097 for VOC 2007), indicating insufficient grouping granularity to generate effective adversarial perturbations. Conversely, increasing the group number beyond 50 (e.g., 75, 100, 150, 200) leads to a continuous decline in ASR (from 80\% to 66\% on VOC 2007, 88\% to 79\% on VOC 2012) while perturbation intensity consistently rises.

For the ML-LIW model, the optimal group number varies slightly across datasets but still centers around 50 groups: MLA-CoE/50 achieves the highest ASR on VOC 2007 (66\%), while MLA-CoE/25 outperforms other settings on VOC 2012 (76\%). Notably, ML-LIW shows more sensitivity to excessive grouping—when the group number exceeds 50, ASR drops sharply (from 62\% to 42\% on VOC 2007, 67\% to 53\% on VOC 2012) accompanied by a significant increase in perturbation metrics. 

Overall, 50 groups were determined to be the optimal setting for MLA-CoE in most experimental settings because it has the highest attack success rate while maintaining a moderate perturbation strength.

\begin{table*}[ht]
	\centering
	\caption{Performance Comparison of MLA-CoE Under Different Group Numbers.}
	\resizebox{\linewidth}{!}{
		\begin{tabular}{c|c|c|c|c|c|c|c|c}
			\hline
			\textbf{Model} & \textbf{Datasets} & \textbf{Group Number} & \textbf{ASR $\uparrow$} & \textbf{$L_1$ Norm $\downarrow$} & \textbf{$L_2$ Norm $\downarrow$} & \textbf{$L_\infty$ Norm $\downarrow$} & \textbf{RSMD $\downarrow$} & \textbf{Mean $\downarrow$} \\
			\hline
			\multirow{12}{*}{ML-GCN} & \multirow{6}{*}{VOC 2007} 
			& 25  & 78\% & \textbf{1728.9169} & \textbf{14.0097} & \textbf{0.5098} & \textbf{2.3020} & \textbf{0.0029} \\
			&                         & 50  & \textbf{81\%} & 2143.8741 & 15.4785 & 0.5200 & 2.5433 & 0.0036 \\
			&                         & 75  & 80\% & 2269.5204 & 16.1417 & 0.5287 & 2.6523 & 0.0038 \\
			&                         & 100 & 76\% & 2473.2510 & 16.9580 & 0.5267 & 2.7864 & 0.0041 \\
			&                         & 150 & 71\% & 2785.4876 & 18.6545 & 0.5190 & 3.0652 & 0.0046 \\
			&                         & 200 & 66\% & 2896.3746 & 18.8096 & 0.5175 & 3.0907 & 0.0048 \\
			\cline{2-9}
			& \multirow{6}{*}{VOC 2012} 
			& 25  & 89\% & \textbf{1354.1373} & \textbf{11.4621} & \textbf{0.4341} & \textbf{1.8834} & \textbf{0.0022} \\
			&                         & 50  & \textbf{91\%} & 1736.1275 & 13.4997 & 0.4593 & 2.2182 & 0.0029 \\
			&                         & 75  & 88\% & 2064.5396 & 14.4705 & 0.4633 & 2.3777 & 0.0034 \\
			&                         & 100 & 88\% & 2265.1478 & 15.5151 & 0.4692 & 2.5493 & 0.0038 \\
			&                         & 150 & 83\% & 2478.2886 & 15.9677 & 0.4461 & 2.6237 & 0.0041 \\
			&                         & 200 & 79\% & 2828.9762 & 17.5272 & 0.4492 & 2.8800 & 0.0047 \\
			\hline
			\multirow{12}{*}{ML-LIW} & \multirow{6}{*}{VOC 2007} 
			& 25  & 64\% & \textbf{2412.3832} & \textbf{17.3988} & \textbf{0.4959} & \textbf{2.8589} & \textbf{0.0040} \\
			&                         & 50  & \textbf{66\%} & 2764.6858 & 18.6392 & 0.5232 & 3.0627 & 0.0046 \\
			&                         & 75  & 62\% & 2968.3515 & 18.9512 & 0.5183 & 3.1139 & 0.0049 \\
			&                         & 100 & 57\% & 3284.4308 & 19.8273 & 0.4934 & 3.2579 & 0.0055 \\
			&                         & 150 & 51\% & 3510.1739 & 19.6399 & 0.4928 & 3.2271 & 0.0058 \\
			&                         & 200 & 42\% & 3351.0727 & 18.7688 & 0.4585 & 3.0839 & 0.0056 \\
			\cline{2-9}
			& \multirow{6}{*}{VOC 2012} 
			& 25  & \textbf{76\%} & \textbf{2673.1896} & 18.5647 & \textbf{0.5164} & 3.0504& \textbf{0.0044} \\
			&                         & 50  & 73\% & 2907.0594 & \textbf{18.2506} & 0.5165 & \textbf{2.9988} & 0.0048 \\
			&                         & 75  & 67\% & 2991.4873 & 18.3562 & 0.4945 & 3.0162 & 0.0050 \\
			&                         & 100 & 71\% & 3709.3298 & 20.4307 & 0.5086 & 3.3570 & 0.0062 \\
			&                         & 150 & 64\% & 3760.3481 & 20.6627 & 0.4999 & 3.3951 & 0.0062 \\
			&                         & 200 & 53\% & 3920.0756 & 20.8905 & 0.4881 & 3.4326 & 0.0065 \\
			\hline
		\end{tabular}
	}
	\label{tab:ablation_group}
\end{table*}

\begin{figure}[t]
	\centering
	\includegraphics[width=0.45\textwidth]{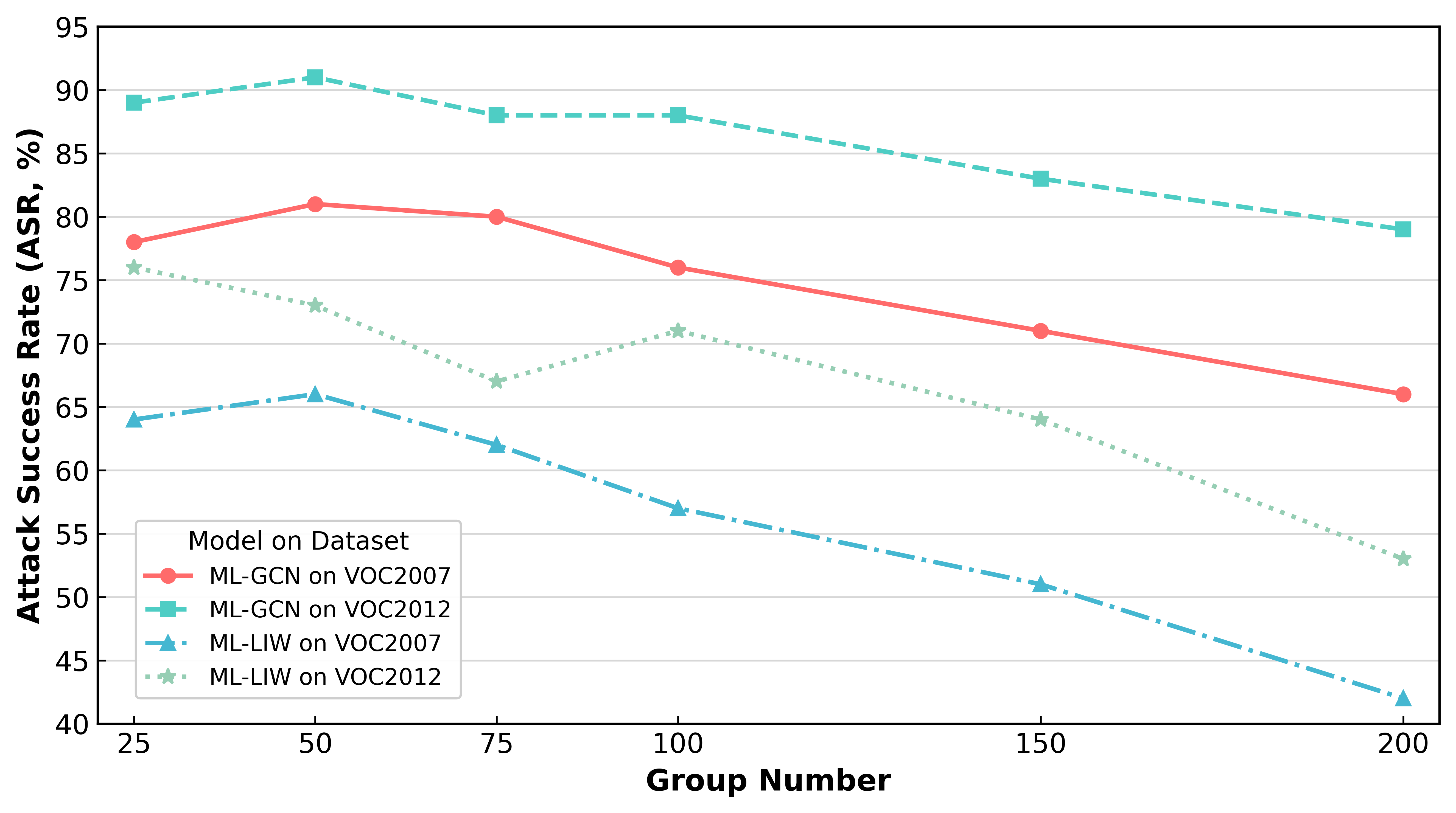}  
	\caption{ASR variation with group number under different models and datasets.}
	\label{fig:asr_vs_groups}
\end{figure}

As illustrated in Fig \ref{fig:asr_vs_groups}, we present the variation of ASR with respect to the group number across four experimental settings.
All curves exhibit a similar trend, where the ASR first rises to a peak at 50 groups and then gradually decreases as the group number continues to increase.
This observation confirms that an appropriate group number effectively balances attack performance, while excessive grouping leads to a decline in attack effectiveness.

\subsection{Ablation Study on Fixed Grouping Strategies}
To investigate the impact of group adjustment mechanisms on MLA-CoE performance, we conduct ablation experiments comparing two distinct strategies: fixed grouping and incremental grouping. 

These two strategies correspond to the dynamic group update mechanism in Line 17 of Algorithm~\ref{alg:MLA-CoE} ($N = N + N_{\delta}$). Specifically:
\begin{itemize}
	\item \textbf{Incremental Grouping:} The group number increases dynamically during optimization by setting $N_{\delta} > 0$ (e.g., $N_{\delta} = 10$). After traversing all current groups, the group count increases by $N_{\delta}$, allowing the algorithm to refine optimization with finer granularity as the attack progresses.
	\item \textbf{Fixed Grouping:} The group number remains constant throughout the attack by setting $N_{\delta} = 0$. The algorithm maintains the initial group number without dynamic subdivision.
\end{itemize}
In this experiment, the incremental grouping strategy initializes with base group numbers (25, 50, 75, 100) and increases the group count by $N_{\delta} = 10$ after each full traversal. The fixed grouping strategy maintains constant group numbers (25, 50, 75, 100) throughout adversarial generation. For clear distinction, methods using fixed grouping are named as \textbf{MLA-CoE/$N$ F} in Table~\ref{tab:ablation_segment_new}, where $N$ denotes the group number and \textbf{F} indicates fixed grouping.

Experimental results in Table~\ref{tab:ablation_segment_new} demonstrate that the incremental grouping strategy consistently achieves better performance than the fixed grouping strategy under all comparable settings. By dynamically adjusting the number of groups during the attack process, the incremental mechanism effectively improves the stability and robustness of MLA-CoE across different initial group numbers. 

In particular, when the attack starts with a relatively small number of groups (e.g., 25 or 50 groups), the advantages of incremental grouping become more significant: it avoids the performance degradation caused by insufficient segmentation granularity in the fixed grouping strategy, and enables the algorithm to maintain a more stable and reliable attack performance. For example, comparing MLA-CoE/50 (incremental) in Table~\ref{tab:ablation_group} with MLA-CoE/50 F (fixed) in Table~\ref{tab:ablation_segment_new}, the incremental version achieves higher ASR on both ML-GCN VOC 2007 (81\% vs. 80\%) and ML-LIW VOC 2012 (73\% vs. 65\%).
\begin{table*}[ht]
	\centering
	\caption{Performance of MLA-CoE With the Fixed Grouping Strategy ($N_{\delta} = 0$).}
	\resizebox{\linewidth}{!}{
		\begin{tabular}{c|c|c|c|c|c|c|c|c}
			\hline
			\textbf{Model} & \textbf{Datasets} & \textbf{Method} & \textbf{ASR $\uparrow$} & \textbf{$L_1$ Norm $\downarrow$} & \textbf{$L_2$ Norm $\downarrow$} & \textbf{$L_\infty$ Norm $\downarrow$} & \textbf{RSMD $\downarrow$} & \textbf{Mean $\downarrow$} \\
			\hline
			\multirow{8}{*}{ML-GCN} & \multirow{4}{*}{VOC 2007} 
			& MLA-CoE/25 F  & 77\% & 2047.7306 & 15.1736 & \textbf{0.5039} & 2.4932 & 0.0034 \\
			&                         & MLA-CoE/50 F  & \textbf{80\%} & \textbf{1969.2322} & \textbf{14.6374} & 0.5142 & \textbf{2.4051} & \textbf{0.0033} \\
			&                         & MLA-CoE/75 F  & \textbf{80\%} & 2294.1002 & 16.0882 & 0.5184 & 2.6435 & 0.0038 \\
			&                         & MLA-CoE/100 F & 78\% & 2479.5776 & 16.7873 & 0.5241 & 2.7584 & 0.0041 \\
			\cline{2-9}
			& \multirow{4}{*}{VOC 2012} 
			& MLA-CoE/25 F  & 88\% & \textbf{1667.3269} & \textbf{12.4189} & \textbf{0.4524} & \textbf{2.0406} & \textbf{0.0028} \\
			&                         & MLA-CoE/50 F  & \textbf{91\%} & 1749.3422 & 13.0597 & 0.4883 & 2.1459 & 0.0029 \\
			&                         & MLA-CoE/75 F  & \textbf{91\%} & 1972.6400 & 14.3178 & 0.4818 & 2.3526 & 0.0033 \\
			&                         & MLA-CoE/100 F & 89\% & 2491.2755 & 16.2244 & 0.4791 & 2.6659 & 0.0041 \\
			\hline
			\multirow{8}{*}{ML-LIW} & \multirow{4}{*}{VOC 2007} 
			& MLA-CoE/25 F  & 62\% & 2693.5774 & 18.6351 & 0.5109 & 3.0620 & 0.0045 \\
			&                         & MLA-CoE/50 F  & 61\% & \textbf{2531.7895} & \textbf{17.4719} & \textbf{0.4969} & \textbf{2.8709} & \textbf{0.0042} \\
			&                         & MLA-CoE/75 F  & \textbf{63\%} & 3104.2051 & 18.6850 & 0.4829 & 3.0702 & 0.0052 \\
			&                         & MLA-CoE/100 F & 59\% & 3340.2728 & 19.5461 & 0.4845 & 3.2117 & 0.0055 \\
			\cline{2-9}
			& \multirow{4}{*}{VOC 2012} 
			& MLA-CoE/25 F  & 60\% & 2931.3785 & 18.6991 & \textbf{0.4984} & 3.0725 & 0.0049 \\
			&                         & MLA-CoE/50 F  & 65\% & \textbf{2766.9588} & \textbf{18.1027} & 0.5156 & \textbf{2.9745} & \textbf{0.0046} \\
			&                         & MLA-CoE/75 F  & \textbf{69\%} & 3224.9207 & 19.3945 & 0.5088 & 3.1868 & 0.0054 \\
			&                         & MLA-CoE/100 F & 65\% & 3572.5724 & 20.1827 & 0.5080 & 3.3163 & 0.0059 \\
			\hline
		\end{tabular}
	}
	\label{tab:ablation_segment_new}
\end{table*}

\subsection{Ablation Study on Sparsification Strategy}
To validate the effectiveness of the sparsification strategy, we conduct ablation experiments to compare the performance of MLA-CoE with and without sparsification under different group numbers (25, 50, 75, and 100). 
The sparsification strategy is designed to filter out redundant non-zero pixels in the initial perturbation: each non-zero pixel is randomly set to 0 with a probability of 0.75, retaining only the critical pixels that contribute to adversarial effectiveness for subsequent optimization. This design aims to reduce query counts, accelerate algorithm convergence, and ultimately achieve better attack performance.

It is important to note that, for ease of distinction and clarity, all methods in Table~\ref{tab:ablation_segment_no_random} are named in the form of \textbf{MLA-CoE/$N$ w/o S}, where $N$ represents the group number and \textbf{w/o S} is short for \textbf{without sparsification}.

As clearly illustrated in Table~\ref{tab:ablation_segment_no_random} (MLA-CoE without sparsification), Table~\ref{tab:ablation_group} (MLA-CoE with sparsification), and the corresponding experimental figures, the sparsification strategy significantly outperforms the non-sparsified scheme in almost all experimental settings, delivering higher attack success rates (ASR) and more favorable perturbation metrics. 
Under the same group number, the sparsified MLA-CoE consistently achieves a higher or comparable ASR while maintaining lower perturbation intensity—including smaller $L_1$ norm, $L_2$ norm, RSMD, and perturbation mean—compared to the non-sparsified version (denoted as MLA-CoE/$N$ w/o S in Table~\ref{tab:ablation_segment_no_random}). 
Notably, the performance advantage of sparsification is more prominent at 50 and 75 groups: for ML-GCN on VOC 2007, the sparsified scheme achieves an ASR of 81\% at 50 groups, which is 3\% higher than the non-sparsified scheme (78\%, denoted as MLA-CoE/50 w/o S); for ML-GCN on VOC 2012, the sparsified scheme reaches an ASR of 91\% at 50 groups, outperforming the non-sparsified scheme (87\%, denoted as MLA-CoE/50 w/o S) by 4\%. Even at 25 and 100 groups, the sparsified scheme either matches or exceeds the ASR of the non-sparsified scheme (MLA-CoE/25 w/o S and MLA-CoE/100 w/o S) while reducing perturbation, fully demonstrating its superiority.

\begin{figure}[ht]
	\centering
	\includegraphics[width=1.0\linewidth]{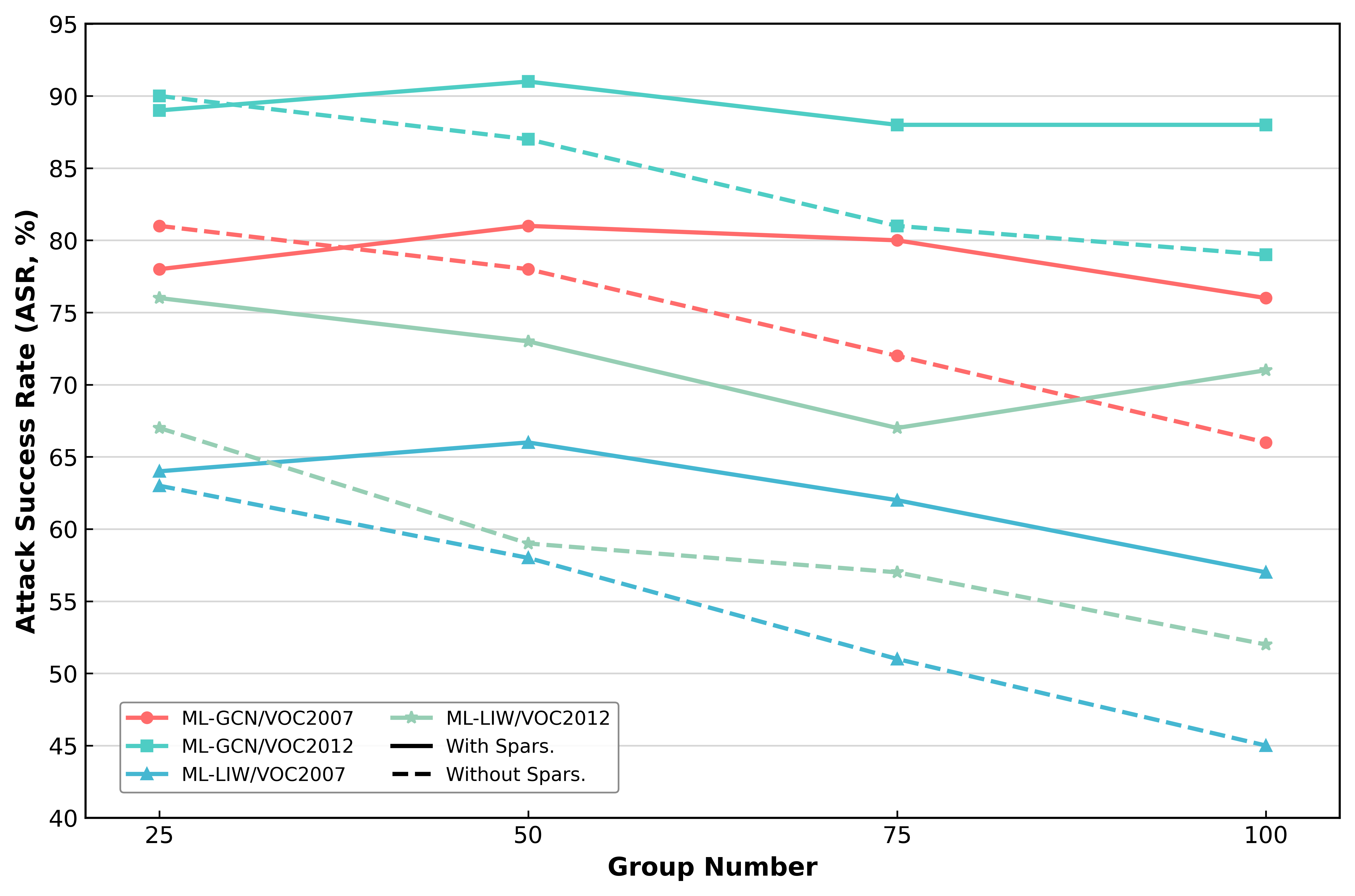}
	\caption{Attack success rate (ASR) of MLA-CoE with and without sparsification under different group numbers. (Note: The non-sparsified version corresponds to MLA-CoE/$N$ w/o S in Table~\ref{tab:ablation_segment_no_random}.)}
	\label{fig:asr_sparse}
\end{figure}

\begin{figure}[ht]
	\centering
	\includegraphics[width=1.0\linewidth]{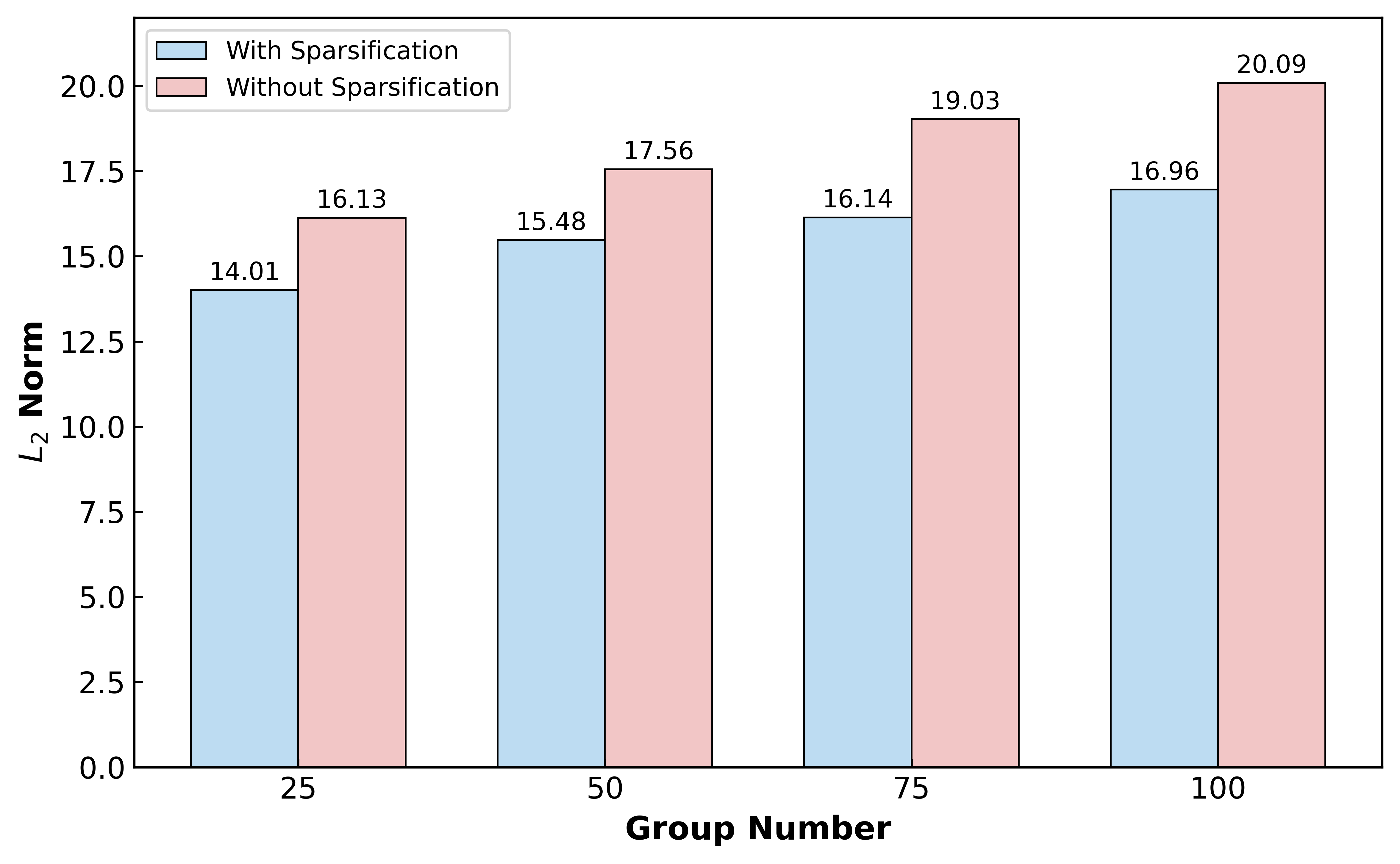}
	\caption{$L_2$ norm comparison of MLA-CoE with and without sparsification on ML-GCN-VOC 2007 under different group numbers. (Note: The non-sparsified version corresponds to MLA-CoE/$N$ w/o S in Table~\ref{tab:ablation_segment_no_random}.)}
	\label{fig:l2_sparse}
\end{figure}

This superiority is further visualized in Fig.~\ref{fig:asr_sparse}, which intuitively presents the ASR comparison between MLA-CoE with and without sparsification under different group numbers. 
Fig.~\ref{fig:asr_sparse} clearly shows that in all control experiments, the ASR of the sparsified MLA-CoE is higher than the non-sparse version in most cases (MLA-CoE/$N$ w/o S), and this difference becomes more pronounced as the number of groups increases.
In terms of perturbation magnitude, Fig.~\ref{fig:l2_sparse} takes ML-GCN on VOC 2007 as an example to show the $L_2$ norm comparison between the two schemes. 
The figure clearly indicates that the sparsified MLA-CoE maintains a significantly lower $L_2$ norm than the non-sparsified version (MLA-CoE/$N$ w/o S) under the same group number, confirming that the sparsification strategy can effectively reduce perturbation while improving ASR.

Experimental results firmly confirm that the proposed sparsification strategy is highly effective and recommended. By eliminating redundant pixels and focusing optimization on critical regions, sparsification not only reduces the number of queries and accelerates algorithm convergence but also significantly improves attack performance, achieving higher ASR and more imperceptible perturbations. Regardless of the group number (25, 50, 75, or 100), the sparsified MLA-CoE consistently delivers better experimental results than its non-sparsified counterpart (MLA-CoE/$N$ w/o S).

\begin{table*}[ht]
	\centering
	\caption{Performance of MLA-CoE Without the Sparsification Strategy.}
	\resizebox{\linewidth}{!}{
		\begin{tabular}{c|c|c|c|c|c|c|c|c}
			\hline
			\textbf{Model} & \textbf{Datasets} & \textbf{Method} & \textbf{ASR $\uparrow$} & \textbf{$L_1$ Norm $\downarrow$} & \textbf{$L_2$ Norm $\downarrow$} & \textbf{$L_\infty$ Norm $\downarrow$} & \textbf{RSMD $\downarrow$} & \textbf{Mean $\downarrow$} \\
			\hline
			\multirow{8}{*}{ML-GCN} & \multirow{4}{*}{VOC 2007}
			& MLA-CoE/25 w/o S & \textbf{81\%} & \textbf{2227.6534} & \textbf{16.1322} & \textbf{0.5080} & \textbf{2.6507} & \textbf{0.0037} \\
			& & MLA-CoE/50 w/o S & 78\% & 2783.0423 & 17.5558 & 0.5112 & 2.8847 & 0.0046 \\
			& & MLA-CoE/75 w/o S & 72\% & 3304.3550 & 19.0265 & 0.5080 & 3.1263 & 0.0055 \\
			& & MLA-CoE/100 w/o S & 66\% & 3535.1987 & 20.0889 & 0.5200 & 3.3009 & 0.0059 \\
			\cline{2-9}
			& \multirow{4}{*}{VOC 2012}
			& MLA-CoE/25 w/o S & \textbf{90\%} & \textbf{1953.0113} & \textbf{13.9744} & \textbf{0.4424} & \textbf{2.2962} & \textbf{0.0032} \\
			& & MLA-CoE/50 w/o S & 87\% & 2479.0166 & 15.9040 & 0.4618 & 2.6132 & 0.0041 \\
			& & MLA-CoE/75 w/o S & 81\% & 2875.4538 & 16.6736 & 0.4490 & 2.7397 & 0.0048 \\
			& & MLA-CoE/100 w/o S & 79\% & 3288.4048 & 18.3211 & 0.4424 & 3.0104 & 0.0055 \\
			\hline
			\multirow{8}{*}{ML-LIW} & \multirow{4}{*}{VOC 2007}
			& MLA-CoE/25 w/o S & \textbf{63\%} & \textbf{3129.3296} & \textbf{19.0787} & \textbf{0.4581} & \textbf{3.1349} & \textbf{0.0052} \\
			& & MLA-CoE/50 w/o S & 58\% & 3772.0207 & 20.5392 & 0.4884 & 3.3749 & 0.0063 \\
			& & MLA-CoE/75 w/o S & 51\% & 4172.6063 & 20.9322 & 0.4778 & 3.4394 & 0.0069 \\
			& & MLA-CoE/100 w/o S & 45\% & 4277.7578 & 21.1132 & 0.4581 & 3.4692 & 0.0071 \\
			\cline{2-9}
			& \multirow{4}{*}{VOC 2012}
			& MLA-CoE/25 w/o S & \textbf{67\%} & \textbf{3009.5843} & \textbf{18.9483} & \textbf{0.4937} & \textbf{3.1134} & \textbf{0.0050} \\
			& & MLA-CoE/50 w/o S & 59\% & 3312.5319 & 19.5952 & 0.5107 & 3.2197 & 0.0055 \\
			& & MLA-CoE/75 w/o S & 57\% & 3877.3934 & 20.7098 & 0.5261 & 3.4029 & 0.0064 \\
			& & MLA-CoE/100 w/o S & 52\% & 4085.0981 & 20.9495 & 0.4937 & 3.4423 & 0.0068 \\
			\hline
		\end{tabular}
	}
	\label{tab:ablation_segment_no_random}
\end{table*}

\section{Conclusion}
\label{sec6}
This paper addresses the challenge of multi-label black-box attacks, where existing evolutionary-based methods suffer from high query costs and slow convergence in large-scale optimization. We propose MLA-CoE, a cooperative co-evolutionary framework enhanced with a sparsification strategy. MLA-CoE first reduces the search space through sparsification, then decomposes the complex adversarial generation problem into lower-dimensional subproblems via cooperative co-evolution. Each subproblem is independently optimized within its subpopulation and collaboratively evaluated to ensure coordination among groups. This mechanism effectively balances perturbation minimization with attack effectiveness.
Extensive experiments on ML-GCN and ML-LIW models demonstrate that MLA-CoE significantly outperforms the mainstream algorithms ML-LBA and ML-BA in terms of attack success rate, query efficiency, and perturbation stealthiness. These results validate the effectiveness of cooperative co-evolution in multi-label adversarial attacks.

For future work, we plan to explore adaptive sparsification and dynamic grouping strategies to further enhance query efficiency, making the attack algorithm more intelligent and practical for real-world multi-label learning systems.

\bibliographystyle{IEEEtran}
\bibliography{ref}

\end{document}